\documentclass{article}

   \usepackage[nonatbib]{neurips_data_2023}

\usepackage{etoolbox}

\usepackage[utf8]{inputenc} 
\usepackage[T1]{fontenc}    
\usepackage{hyperref}       
\usepackage{url}            
\usepackage{booktabs}       
\usepackage{amsfonts}       
\usepackage{nicefrac}       
\usepackage{microtype}      
\usepackage[dvipsnames]{xcolor}         

\usepackage{graphicx} 

\usepackage{makecell}
\usepackage{amsfonts}
\usepackage{amsmath}
\usepackage{amsthm}
\usepackage{subfigure}
\usepackage{multirow}
\usepackage{stfloats}
\usepackage{hyperref}
\usepackage{lipsum}
\usepackage{algorithm}
\usepackage{algpseudocode}
\usepackage{hyperref}
\hypersetup{
    colorlinks=true,
    linkcolor=blue,
    filecolor=magenta,      
    urlcolor=black,
    pdftitle={Overleaf Example},
    pdfpagemode=FullScreen,
    }
\usepackage{enumitem}

\DeclareMathOperator*{\argmax}{arg\,max}

\newtheorem{lemma}{Lemma}
\newtheorem{theorem}{Theorem}

\title{Establishing Reliability Metrics for Reward Models in Large Language Models}
\author{Anonymous}
\date{April 2024}

\author{%
  Yizhou Chen, Yawen Liu, Xuesi Wang\\ \textbf{Qingtao Yu, Guangda Huzhang}\\
  Shopee Pte. Ltd.\\
  \And
  Anxiang Zeng, Han Yu \\
  Nanyang Technological University \\
  \And
  Zhiming Zhou \\
  Shanghai University of Finance and Economics \\
}

\begin{document}

\maketitle

\begin{abstract}

The reward model (RM) that represents human preferences plays a crucial role in optimizing the outputs of large language models (LLMs), e.g., through reinforcement learning from human feedback (RLHF) or rejection sampling. However, a long challenge for RM is its uncertain reliability, i.e., LLM outputs with higher rewards may not align with actual human preferences.
Currently, there is a lack of a convincing metric to quantify the reliability of RMs.
To bridge this gap, we propose the \textit{\underline{R}eliable at \underline{$\eta$}} (RETA) metric, which directly measures the reliability of an RM by evaluating the average quality (scored by an oracle) of the top $\eta$ quantile responses assessed by an RM. On top of RETA, we present an integrated benchmarking pipeline that allows anyone to evaluate their own RM without incurring additional Oracle labeling costs. Extensive experimental studies demonstrate the superior stability of RETA metric, 
providing solid evaluations of the reliability of various publicly available and proprietary RMs. When dealing with an unreliable RM, we can use the RETA metric to identify the optimal quantile from which to select the responses.

\end{abstract}


\section{Introduction}

Large language models (LLMs) such as GPT-4 \cite{gpt4}, Claude 3 \cite{claude3}, and Gemini \cite{gemini} have made profound impacts on both the AI community and society at large. The advancement of these models can be attributed to a systematic, tripartite training approach. 
After pre-training \cite{radford2018improving,bert,brown2020language} and supervised fine-tuning (SFT) \cite{wei2021finetuned,alpaca}, the final phase involves human preference alignment, often achieved through
{reinforcement learning from human feedback} (RLHF) \cite{ouyang2022training,bai2022training,christiano2017deep}, which fine-tunes model outputs through a proxy \emph{reward model} (RM) to align with human values and preferences. Such RMs are developed and trained using datasets with binary preferences that mimic human judgment.

The improvement of LLM capabilities through human preference alignment relies on the quality of the RMs. However, the reliability of the RM is a long challenge \cite{llama2,casper2023open}: 
When there are ``loopholes'' in the RM, i.e., specific inputs that can trigger higher reward but do not align with actual human preferences \cite{warm}, the LLM output is prone to reward hacking (a.k.a. overoptimization) \cite{gao2023scaling,mckinney2023fragility,rewardmisspecification1}. In RLHF, this often results in policy model performance decline (manifested in the forms of linguistically flawed and excessively verbose outputs, biased responses, or the generation of unsafe contents \cite{lengthcorrelation,sycophancy,hendrycks2022x,warm}). 
These issues can also occur when the RM is not invoked by RLHF but is directly employed as a ranking mechanism for selecting responses \cite{gao2023scaling,webgpt}.
Unreliable RMs cannot act as dependable policy model selectors in RLHF, increasing the likelihood of failures in the re-training of policy models \cite{rewardmisspecification1,mckinney2023fragility}.

The RM reliability issue (i.e., loopholes) is an inherent consequence of the limited scope of the training data distribution, arising from reward misspecification between the proxy RM and actual human preferences \cite{rewardmisspecification1,rewardmisspecification2,warm}. 
While the prevailing approaches concentrate on circumventing these loopholes by applying KL regularization techniques to the policy model (e.g., during proximal policy optimization (PPO) in RLHF \cite{ouyang2022training,bai2022training}), works on diminishing such loopholes (e.g., through ensembling or pooling of RMs \cite{ensemble1,ensemble2,ensemble3,warm}) have emerged. 
Nevertheless, currently there are few direct assessments of RM reliability. This may be attributed to the prevailing emphasis on the resulting policy model and the shared assumption that the efficacy and reliability of the RM can be assessed indirectly through the quality of the policy it generates via PPO.

However, indirectly assessing the reliability of RM through the resulting policy model has three major drawbacks. 
Firstly, it requires RLHF training (e.g., PPO) to be completed as a prerequisite. This entails extensive hyperparameter tuning, which is time-consuming and costly, particularly when a large number of candidate RMs are involved. 
Secondly, a single attempt of RLHF training is not representative, and the outcome may not generalize across different architectures of the policy model \cite{rewardmisspecification1,mckinney2023fragility,gao2023scaling}. 
Thirdly, RLHF often includes regularization to circumvent the unreliability issue of RMs. This can compromise the validity of using final policy model performance as a proxy for RM evaluation, since the efficacy of the regularization techniques adopted can be a confounding factor of the reliability of RM.
Given these issues, there is a clear and pressing need for methods that can directly assess the reliability of RMs.


The \textit{best-of-$n$} (BON) sampling, also referred to as rejection sampling or re-ranking, has emerged as a compelling alternative for optimizing responses against the reward model. 
In BON sampling, $n$ responses are first collected and the RM is used to pick the one with the highest proxy RM score.
BON sampling does not necessitate further RL training, despite incurring the expense of more inference-time computing.
It has proven to be competitive and can sometimes surpass the performance of RL \cite{webgpt,gao2023scaling}.
BON sampling can also be extended into a plot of {BON curve} (the oracle score of the BON sample versus $n$), which is reckoned as an RM evaluation approach. BON curve assumes the availability of an oracle (e.g., a human) to provide labels. These labels are utilized to assess the quality of responses selected by the proxy RM.
However, we found that using the BON curve as a direct RM evaluation method faces two limitations.
Firstly, BON curve exhibits substantial variance concerning the selected ranking, i.e., selecting the best response versus the second-best response yields significantly different evaluation results. This is due to its reliance on only a single response according to the RM.
Secondly, the evaluation results might be sensitive to the selection of prompts used to construct the benchmark.

\begin{figure}[t]
    \centering
    \includegraphics[width=0.99\textwidth]{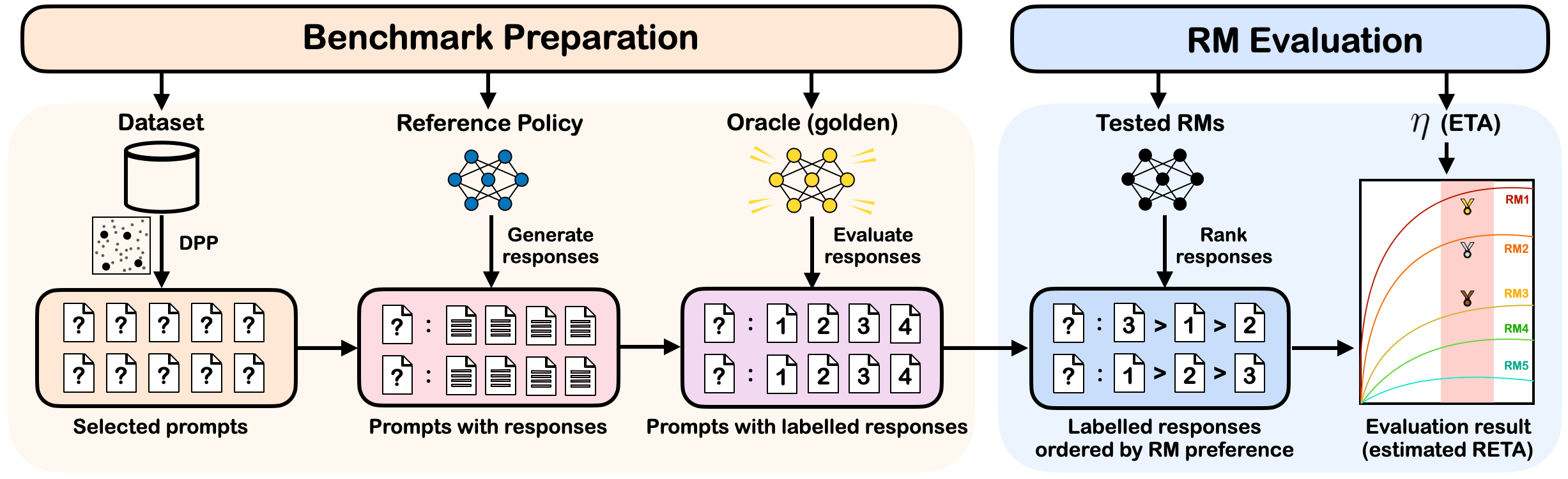}
    \vspace{-2mm}
    \caption{The benchmark building pipeline and the computation of the RETA metric.}
    \label{fig:RETA flow}
\end{figure}

In this regard, we introduce \textit{\underline{R}eliability at \underline{$\eta$}} (RETA).
RETA is a scalar metric that evaluates the average quality across the highest $\eta$ quantile of responses, rather than relying on a single selected response. 
A theoretical investigation reveals that RETA offers clear statistical meaning, and exhibits favorable convergence properties.
In addition, we apply normalization to RETA by dividing it by the average score of the responses (per prompt), enabling it to have a meaningful comparative interpretation. This normalization also enhances the metric's robustness against variations in the selection of prompt subsets. 
The RETA metrics at various values of $\eta$ can be consolidated to form a \textbf{RETA curve} (the metric value versus different $\eta$), allows easy visualization of rich information about RM reliability across every preference distributional quantile.
All the above benefits are attained while maintaining a manageable oracle labeling cost during benchmark preparation, and there is no labeling cost incurred for any round of RM evaluation.

In order to streamline the computation of RETA benchmarking, we have developed an integrated benchmarking pipeline that includes guidelines for dataset curation and a specifically designed procedure for metric estimation. 
We illustrate the complete flow in Figure~\ref{fig:RETA flow}. 
Our dataset curation utilizes {determinantal point process} (DPP) sampling to select prompts that are both diverse and representative. Subsequently, a reference policy is employed to generate candidate responses for labeling by the oracle.
As for metric computation, we employ an asymptotic unbiased estimation to ensure consistent evaluation, and utilize resampling techniques to empirically minimize the potential bias and variance impact.
This comprehensive pipeline is beneficial for individuals seeking to evaluate the reliability of RMs based on specific criteria (e.g., helpfulness or harmlessness).

Our contributions are summarized as follows:
\vspace{-2mm}
\begin{itemize}[leftmargin=10pt]
\setlength\itemsep{-0.1em}
    \item We introduce RETA for assessing the reliability of RMs, demonstrating that it offers clear statistical meaning, favorable convergence properties, robustness to variations in the selection of prompts, all while maintaining a manageable oracle labeling cost.
    \item We provide an integrated benchmarking pipeline. We made the entire code available to facilitate further community research, and also released our benchmark to ensure high reproducibility.
    \item We benchmarked a variety of reward models and found that the RETA metric consistently delivers useful results, offering advantages over other candidate metrics evaluated. 
\end{itemize}


\section{Related Works}
Existing LLM benchmarks such as AlpacaEval~\cite{alpaca} and MTBench~\cite{mtbench} are predominantly designed for evaluating policy models rather than RMs. Some studies on assessing RMs focus on test accuracy performance or granular test accuracy performance~\cite{llama2}. In this context, the only solid benchmark is RewardBench~\cite{rewardbench}. However, our analysis indicates that accuracy-based metrics / benchmarks do not necessarily reflect reliability.

The current research lacks focus on establishing a reliability benchmark for RMs. Studies have suggested methods to address RM over-optimization by utilizing KL regularization techniques in the policy model (e.g., during PPO in RLHF \cite{ouyang2022training, bai2022training}), or by reducing vulnerabilities through ensembling / pooling of RMs \cite{ensemble1, ensemble2, ensemble3, warm}. However, these approaches have not established benchmarks to directly assess the reliability of RM.

The BON curve \cite{webgpt,gao2023scaling} is a promising metric for assessing reliability. However, no benchmark has yet been established based on this approach. 
Our study introduces the RETA metric to improve the BON-based metric to construct a much needed RM reliability evaluation benchmark for LLMs.

\section{The Proposed RM Reliability Evaluation Metric}
\subsection{Preliminaries}
In order to evaluate the reliability of a reward model, it is necessary to establish a \textit{reference policy} (RP) model that serves as a proxy for generating responses.
The RP encompasses the input distribution for the RM to quantify the region of interest for evaluation. Hence, it should be competent and comprehensive. Firstly, its responses should achieve satisfactory quality to avoid overemphasizing trivial regions with low-quality inputs.
Secondly, its responses should be general and diverse. For this study, we adopt Llama2-7b-Chat \cite{llama2} at temperature $T=1$ as our RP model.\footnote{Additional discussions regarding the selection of the RP model can be found in the Appendix~\ref{appendix:RP model}.}

Furthermore, similar to works like \cite{gao2023scaling,warm} that utilizes BON curve, our proposed approach requires an oracle to assess the response quality. The evaluation can take the form of a direct score, such as a scalar ranging from 1 to 10.
Let ${J}_q:a\to \mathbb{R}^{+}$ denote the \emph{oracle $J$'s evaluation} of response $a$ to prompt $q$.
To mitigate the expense associated with human labeling, as well as to ensure consistent feedback, we utilize the state-of-the-art GPT-4 model\footnote{The specific version of GPT4 we utilized throughout the paper is \href{http://platform.openai.com/docs/models/gpt-4-turbo-and-gpt-4}{\textit{gpt-4-turbo-2024-04-09}}.} as an oracle for evaluating responses. The comprehensive scoring procedure, including our prompt template, is detailed in Appendix~\ref{appendix:oracle}. This process ultimately derives an average from 10 sampled scores produced by GPT-4.

\textbf{BON curve.} We first examine the \emph{best-of-$n$} (\text{BON}) curve, a curve that utilizes BON sampling and can be employed as a means to assess the reliability of an RM.
Let $\mathcal{A}_q=\{a|a\sim\theta_{q}(a)\}$ ($|\mathcal{A}_q|=n$) represent a candidate response set i.i.d sampled (generated) by the RP model $\theta$ for a prompt $q$.
The BON response of an RM $Y$ is its chosen best response $\argmax_{a \in \mathcal{A}_q} Y_{q}(a)$ from the candidate pool $\mathcal{A}_q$.
The BON curve is then defined as a score curve concerning $n$ ($n=0,1,\ldots$).
\begin{equation}
\textstyle
\text{BON}_{Y}(n)=
\frac{1}{|\mathcal{Q}|}\sum_{q\in\mathcal{Q}}\mathbb{E}_{\mathcal{A}_q: |\mathcal{A}_q| = n}
\left[{J}_q\left(\argmax_{a \in \mathcal{A}_q} Y_{q}(a)\right)
\right].
\label{equ:bon}
\end{equation}
Here, $\mathcal{Q}$ represents the set of prompts used for metric evaluation.
The BON curve allows us to visualize the relationship between the best quality of responses and the number of samples. In addition, the KL divergence (between the selected best response from a set of $n$ samples versus a random sample from RP) is theoretically computed as $\text{KL}=\log n -\frac{n-1}{n}$ \cite{stiennon2020learning}. By examining the BON curve versus KL divergence, we can identify the point at which the performance of best sample starts to decline, indicating the onset of potential overfitting. 

Our experiments reveal that BON curves exhibits high variance concerning the selected ranking, making it challenging to derive a stable metric from the curve itself. 
This variance arises because the BON curve focuses on selecting a single response from the candidate pool, which can lead to low stability and might not adequately represent the preferences of the RM model.

\subsection{RETA: Reliability at ETA quantile}
To address this limitation, we propose an alternative approach based on the \textit{\underline{b}est-\underline{$\eta$} quantile-of-$n$} (BETA) subset where $0<\eta<1$, defined as follows:
\begin{equation}
\textstyle
\text{BETA}_{Y}(\mathcal{A}_q,\eta)\equiv
\argmax_{\mathcal{A} \subset \mathcal{A}_q: |\mathcal{A}| =\eta\times |\mathcal{A}_q|} \sum_{a \in \mathcal{A}} Y_{q}(a),
\label{equ:beta}
\end{equation}
It consists of the top ranked responses by RM $Y$.
Without loss of generality, we first assume $\eta\times |\mathcal{A}_q|$ to be an integer.
Our proposed \emph{\underline{r}eliability-at-\underline{$\eta$} quantile} (RETA), can be formally defined:
\begin{equation}
\textstyle
\text{RETA}_{Y}(\eta)=
\lim_{|\mathcal{A}_q|\to\infty}\;\;\frac{1}{|\mathcal{Q}|}\sum_{q\in\mathcal{Q}}\;
{\mathbb{E}_{\mathcal{A}_q}
\left[
\frac{1}{\eta\times |\mathcal{A}_q|}
\sum_{a\in\text{BETA}_{Y}(\mathcal{A}_q,\eta)} {J}_{q}(a)\right]}\;/\;{\mathbb{E}_{a}[{J}_{q}(a)]}.
\label{equ:reta}
\end{equation}
It can be proven that this metric has a limiting value as $|\mathcal{A}_q|=n$ approaches infinity. 
\begin{theorem}
    Let $F$ denote the cumulative distribution function of the random variable $X\equiv Y_{q}(a)$, $a\sim \theta_{q}(a)$, and let $\Theta(\eta)=\inf (x:F(x)\geq 1-\eta)$ define the quantile function. Assume ${J}_{q}$ is bounded. Then, the following holds at all continuity points $\eta$ of $\Theta$.
\begin{equation}
\textstyle
\text{RETA}_{Y}(\eta)=
\frac{1}{|\mathcal{Q}|}\sum_{q\in\mathcal{Q}}
{\mathbb{E}_{a}\left[{
{J}_{q}(a)|{Y_q(a)\geq \Theta(\eta)}}\right]}\;/\;{\mathbb{E}_{a}[{J}_{q}(a)]}
\label{equ:limit}
\end{equation}
\end{theorem}
The complete proof can be found in Appendix~\ref{appendix:limit}. In our definition of RETA, we opt to include the average oracle score $\mathbb{E}_{a}[{J}_{q}(a)]$ in the denominator as a \textit{normalizer}, although the limit still exists under the same condition without it. The normalizer first calibrates the oracle score, allowing the RETA metric to have a comparative meaning, as the metric value of 1 represents the random baseline. Then, it reduces the effect of prompt selection bias, which is illustrated in our experiment (Sec.~\ref{sec:ppl}).
Intuitively, the above limiting value captures the quality of responses obtained by selecting the tested RM's top $\eta$ quantile from the response distribution, relative to the average response quality.

\subsection{Empirical Estimation of RETA} 
\label{subsec: estimation}
In practice, there are two sources of the potential bias and variance when estimating RETA. Firstly, when the estimation is based on a finite sample size $n$, the proof in Appendix~\ref{appendix:limit} reveals that estimation biases arise from substituting $\Theta(\eta)$ with $(\eta\times n)$-th largest datum among $X_1,\ldots,X_n$ (i.e., replacing the true distributional quantile with the sample quantile). Secondly, we can only evaluate the expectation in Eq.~\eqref{equ:reta} through Monte Carlo estimation in practice. This introduces the second source of potentially large deviation when there are only a limited number of samples of $\mathcal{A}_q$.

Suppose our oracle labeling budget is fixed to $N(\geq n)$ per prompt (i.e. label $N$ responses from $a\sim \theta_{q}(a)$). Reducing the first source of bias requires us to evaluate Eq.~\eqref{equ:reta} using as large an $n$ value as possible. However, in a naive case when $n=N$, we only have one random sample of $\mathcal{A}_q$ for the Monte Carlo estimation, and thus can only evaluate the expectation in a one-off fashion (or resampling with exact size $|\mathcal{A}_q|=N$ every time). This can lead to large deviations due to its sample variance in the second source as shown in our experiments. Thus the choice of $n$ is a trade-off between bias and variance of these two sources.

Determining an optimal value of $n$ is challenging, even when the target estimation value is simply the distribution quantile \cite{bootstrap1,bootstrap2,bootstrap3}, not to mention in our compounded metric.
Given this, we consider all $n$ values within the range $[3N^{\frac{2}{3}},5N^{\frac{2}{3}}]$ and average the estimation results to obtain the final estimated RETA. We empirically select the exponent $\frac{2}{3}$ because it has been shown that bootstrapping the variance estimator for the distribution quantile achieves the fastest convergence rate for the relative error with $n\propto N^{2/3}$ \cite{bootstrap1,bootstrap2}. We then empirically select the constant terms.

As a result, $N$ responses are first sampled and labeled in preparation. Subsequently, for each $n$ in the range $[3N^{\frac{2}{3}},5N^{\frac{2}{3}}]$, we use the following asymptotic unbiased estimator\footnote{The evaluation of BON is much simpler than RETA. BON \cite{webgpt} initially proposed sampling $N\geq n$ responses and subsequently adopting a unbiased estimator, detailed in Appendix~\ref{appendix:bon sampler}.}, where the random variable $\mathcal{A}=\{a_{i_1},\ldots,a_{i_n}\}$ ($1\leq i_1,\ldots,i_n\leq N$) contains $n$ randomly resampled responses:
\begin{equation}
\textstyle
\frac{1}{|\mathcal{Q}|}\sum_{q\in\mathcal{Q}}\left[
\frac{N}{\eta\times n}\mathbb{E}_{\mathcal{A}}
\left[\sum_{a\in\text{BETA}_{Y}(\mathcal{A},\eta)} {J}_{q}(a)\right]/{\sum_{a\in\mathcal{A}_q}{J}_{q}(a)}
\right].
\label{equ:approx}
\end{equation}

The asymptotic unbiasedness of the estimation is straightforward because of the linearity of expectation. Practically, we resampled $200$ samples of $\mathcal{A}$ to evaluate the above expectation.
To handle situations where $\eta\times n$ is non-integer, we employ smoothing techniques as described in Appendix~\ref{appendix:limit}. 

\textbf{Dataset Curation with \text{$k$-DPP} Sampling.}
One key factor to effective dataset curation for benchmarking is prompt sampling. We leverage \textit{Determinantal Point Processes} (DPP) for this purpose. 
It can be defined through a positive semi-definite kernel matrix computed using embeddings of the prompts. The quality of these embeddings directly affects the quality of sampling.
While DPP has been used in prompt sampling before \cite{zhang2024dpp}, our design offers significant improvements. In \cite{zhang2024dpp}, only very short prompts ($\leq8$ tokens) are considered. The embedding is constructed with naive concatenation of token embeddings, which is not a principled approach.

We use the embedding model from OpenAI to obtain effective prompt embeddings. The 
\href{http://platform.openai.com/docs/guides/embeddings}{\textit{text-embedding-3-large}} is OpenAI's newest and most effective embedding model, and has been shown to be useful for generating embeddings for various purposes, including diversity measurement.
After obtaining the embeddings of the prompts in the candidate prompt pool $\mathcal{Q}_0$, $\mathbf{E}=[\mathbf{e}_q]_{q\in \mathcal{Q}_0}$, the kernel matrix is constructed by $\lambda\mathbf{E}^\top\mathbf{E}$, where $\lambda$ is a scaling factor used to downscale the eigenvalues of the kernel matrix to be smaller than $1$ as desired by DPP.

We utilize the prevalent $k$-DPP that models only sets of cardinality $|\mathcal{Q}|=k$. Many existing algorithms can approximate $k$-DPP with arbitrary accuracy within acceptable time. For example, the Monte Carlo Markov Chain (MCMC) algorithm \cite{dppmcmc} converges within $O(|\mathcal{Q}_0|k \log \frac{1}{\epsilon})$ time to generate an $\epsilon$-approximate sample of $k$-DPP. When 
$|\mathcal{Q}_0|$ is particularly large, there are also efficient algorithms like \cite{dppm} with only $O(k^2)$ time complexity, provided that an initial core set can be constructed.

\textbf{Overall Benchmarking Pipeline.}
In summary, we first choose the RP model $\theta$ and the pool of candidate prompts $\mathcal{Q}_0$. Then, we follow the procedure (as illustrated in Figure~\ref{fig:RETA flow}):
\begin{itemize}
\vspace{-2mm}
\setlength\itemsep{-0.2em}
    \item Sample representative prompts $\mathcal{Q} \gets k\text{-DPP}(\lambda\mathbf{E}^\top\mathbf{E})$, where $\mathbf{E}=[\mathbf{e}_q]_{q\in \mathcal{Q}_0}$; 
    \item Curate the response set $\mathcal{A}_q \gets \{a|a\sim\theta_{q}(a)\}$ s.t. $|\mathcal{A}_q|=N$, $\forall$ $q\in\mathcal{Q}$; 
    \item Curate the oracle evaluation set $\{{J}_{q}(a) \;|\; a\in \mathcal{A}_q\}\;\;\forall q\in\mathcal{Q}$; 
    \item \textcolor{blue}{The RM $Y$ to be tested to evaluate the response set $\{{Y}_{q}(a) \;|\; a\in \mathcal{A}_q\}\;\;\forall q\in\mathcal{Q}$;} 
    \item \textcolor{blue}{Estimate RETA for $Y$ by averaging Eq.~\eqref{equ:approx} for $n$ in the range of $[3N^{\frac{2}{3}},5N^{\frac{2}{3}}]$.}
\vspace{-2mm}
\end{itemize}

Finally, we summarize the oracle labeling cost\footnote{The cost includes both input and output token price. However, even though we average over 10 outputs, the output cost is negligible compared to the input because we only request a single output token as score each time.} and computational cost on Table~\ref{tab:cost}. The preparation initially invokes the oracle to evaluate all $kN$ sampled responses, resulting in the same total cost as that of BON \cite{webgpt,gao2023scaling}. There are no additional costs during the metric computation, as the tested RM only needs to judge responses from the existing labeled collections.

\begin{table}[ht]
\centering
\caption{The total labeling cost and computational cost. The oracle labeling cost is computed base on \href{http://platform.openai.com/docs/guides/embeddings}{\textit{text-embedding-3-large}} and \href{http://platform.openai.com/docs/models/gpt-4-turbo-and-gpt-4}{\textit{gpt-4-turbo-2024-04-09}} models we utilized. Let $L$ denote the average length of the prompt. The benchmark preparation phase is presented in \textbf{black}, while the evaluation process is presented in \textcolor{blue}{\textbf{blue}}.}
\scalebox{0.85}{
\begin{tabular}{llccc}
\hline
& & Prompt sampling & Response Curation & \textbf{Total Cost}             \\ \hline
&\textbf{Labeling cost} ($10^{-5}\;\$$ (USD))& $0.013\cdot|\mathcal{Q}_0|L$   & $kNL$   & $(0.013\cdot|\mathcal{Q}_0| + kN)L$           \\
&\textbf{DPP sampling} (times)      & $O(|\mathcal{Q}_0|k)$ & - & $O(|\mathcal{Q}_0|k)$ \\
&\textbf{RP response generation} (times) & - & $kN$                & $kN$ \\ \hline
&\textcolor{blue}{\textbf{RM scoring} (times)}        & - & \textcolor{blue}{$kN$}                & \textcolor{blue}{$kN$} \\ \hline
\end{tabular}}
\label{tab:cost}
\end{table}

\section{The Proposed Benchmark and Experimental Studies}
\label{sec:exp}
\textbf{Reliability-on-Helpfulness.}
Helpfulness is a crucial aspect of an AI agent. To build such a benchmark, we need a suitable candidate pool of prompts. Anthropic-HH \cite{bai2022training} is a human preference dataset on dialogues, consisting of two subsets: Helpful and Harmless. We utilize the prompt set from Anthropic-Helpful to better isolate and understand the reliability of response helpfulness principles. Anthropic-Helpful contains various data splits corresponding to different development stages of an AI assistant. We consider the default test split of the base dataset (2,345 preference samples) \cite{bai2022training}. The reason for using the test set is to focus on evaluating RM reliability on unseen prompts.

For extraction of the prompt, we parse only the context before the final  response. While the preference pairs in Anthropic-Helpful share part of the same context, the number of interaction rounds sometimes differs. If a redundant role header exists compared to another record in the preference pair (e.g., having excessive rounds of dialogue), we remove these excessive rounds. We treat the last responses (generated by the RP model) as the content for comparison.
Finally, we sampled $k=100$ prompts, and obtained $N=256$ RP responses for each prompt. The total labeling cost is $\$199.3$. Note that if we use a cheaper oracle, we can significantly reduce the labeling cost. 
For example, the cost would be $\$61.4$ for Claude-3-Sonnet and $\$10.0$ for GPT-3.5-turbo.

\textbf{Reliability-on-Multi-Turn-Conversation.}
We also consolidated an additional reliability benchmark to assess RM performance in multi-turn dialogues. Due to space constraints, the detailed benchmarking procedure and relevant results are provided in Appendix~\ref{sec:benchmark2}.
 
\textbf{Tested RMs.} 
We assess multiple open-source RMs by 
selecting top ranked RMs from the accuracy benchmark RewardBench~\cite{rewardbench}. 
For meaningful comparison, we limit the parameter size of the tested RMs to 7B. We specifically consider RMs that can output a scalar reward value for a response without needing a reference or comparison response, thus excluding models like PairRM or SteamSHP-flan. Our selection includes: \textcolor{YellowOrange}{Starling-7B} (Starling-RM-7B-alpha \cite{starling}) trained from Llama2-7B-Chat; \textcolor{Salmon}{Deberta-0.3B} (reward-model-deberta-v3-large-v2 \cite{deberta}), a small model; \textcolor{CarnationPink}{RAFT-3B} (hh-rlhf-rm-open-llama-3b~\cite{dong2023raft}), a model specifically trained for the Anthropic-Helpful and has very high test accuracy on it; \textcolor{Blue}{Pythia-1.4B} (oasst-rm-2.1-pythia-1.4b-epoch-2.5 \cite{pythia}), a model excels on Anthropic-Harmless and performs adequately on Anthropic-Helpful dataset; \textcolor{SeaGreen}{Ziya-7B} (Ziya-LLaMA-7B-Reward \cite{ziya}), a model with competent performance on MT-bench. In addition, we include a few RMs trained by ourselves: \textcolor{Brown}{RMv1-7B}, \textcolor{olive}{RMv2-7B}, and \textcolor{Green}{RMv3-7B} are three RMs trained based on the Llama2-7B, using the Anthropic-Helpful dataset and feature distinct configurations of learning rates and various other hyperparameters; \textcolor{RoyalBlue}{RM5H-7B} mirrors the hyperparameter settings of RMv3-7B, but it is characterized by a 5-head architecture: the final reward output is the average values from these five heads; \textcolor{Cyan}{RMEns-3x7B} is our implementation of prediction ensembling \cite{ensemble1,ensemble2}, which combines the outputs from 3 of our reward models RMv1, RMv2, and RMv3. Detailed training procedures for all our reward models are documented in the Appendix~\ref{appendix:self train RMS}.


\subsection{Comprehensive Comparison between RETA and BON}

\begin{figure}[ht]
\vspace{-5mm}
    \centering
    \subfigure[\hspace{-7mm}]{\includegraphics[width=0.25\textwidth]{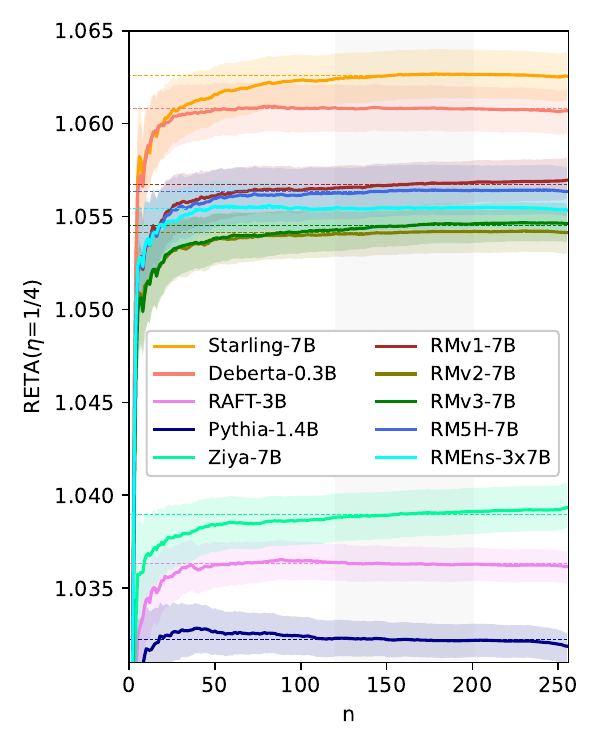}}
    \hspace{-3mm}
    \subfigure[\hspace{-7mm}]{\includegraphics[width=0.25\textwidth]{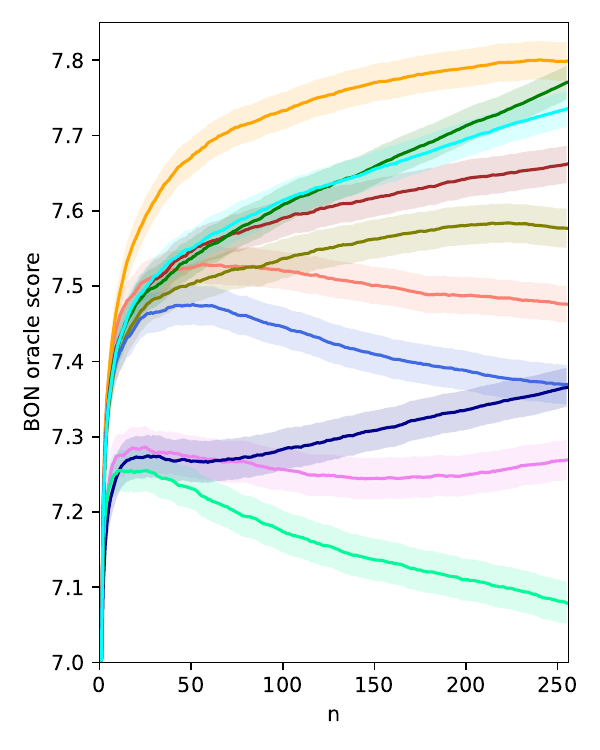}} 
    \hspace{-3mm}
    \subfigure[\hspace{-7mm}]{\includegraphics[width=0.25\textwidth]{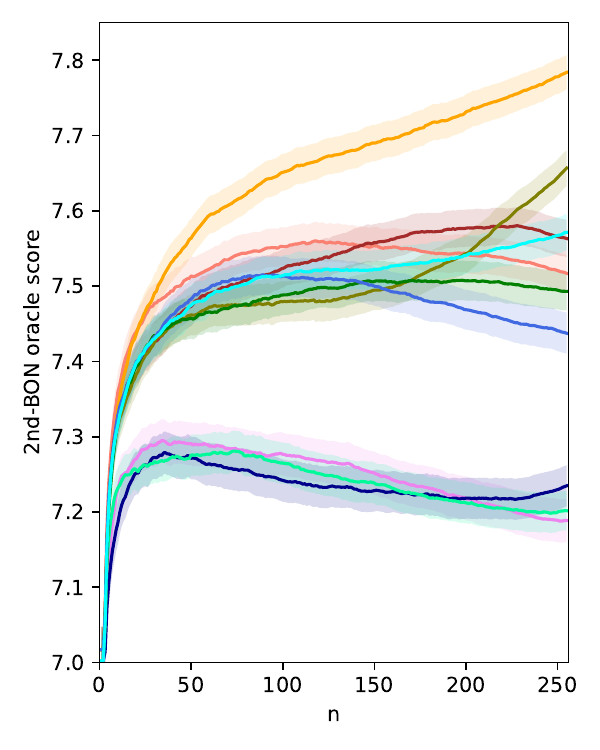}}
    \hspace{-3mm}
    \subfigure[\hspace{-7mm}]{\includegraphics[width=0.25\textwidth]{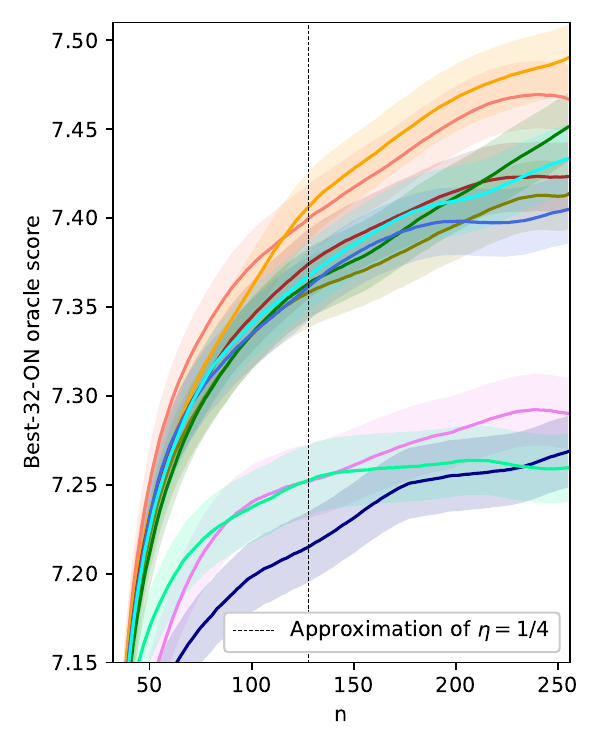}}
    \vspace{-3mm}
    \caption{The results on Reliability-on-Helpfulness benchmark: 
    (a) The estimation of RETA($\eta=1/4$) in Eq.~\eqref{equ:approx} versus resampled size $n$. 
    Dashed horizontal lines mark the limiting values, with the light grey shaded area representing the range used to calculate the final estimation of RETA's limiting values (Sec.~\ref{subsec: estimation}).
    (b) \textit{Best-of-n} (BON) curve versus $n$. The x-axis can also be transformed into KL divergence without loss of generality. 
    (c) \textit{2nd-best-of-n} curve versus $n$.
    (d) The average oracle scores of \textit{best-32-of-n} curve versus $n$. For above figures, the standard error across prompts is plotted.
    } 
    \label{fig:BON vs RETA}
\end{figure}

As depicted in Figure~\ref{fig:BON vs RETA}(a), RETA achieves rapid convergence with respect to $n$ and converges to a desirable limiting value, as predicted in theory. The tail end of the curve ($n>250$) occasionally exhibits instability, corroborating our conjecture that $n\approx N$ might lead to large deviation due to the limited number of Monte Carlo samples of $\mathcal{A}$, underscoring the necessity of the proposed estimation scheme.
A significant disparity in tendencies between BON (Figure~\ref{fig:BON vs RETA}(b)) and RETA$(1/4)$ is clearly visible. 
This disparity naturally raises the question of whose judgment holds greater soundness.

Firstly and most importantly, BON solely focuses on the top response, which by itself can exhibit substantial variance concerning the selected ranking (i.e. selecting the best response versus the second-best response already yields significantly different evaluation results). 
In Figure~\ref{fig:BON vs RETA}(c), it can be observed that the \textit{2nd-best-of-n} curve deviates significantly from the \textit{best-of-n} curve (Figure~\ref{fig:BON vs RETA}(b)).
The counter-intuitive outcome greatly undermines the reliability of the BON curve. The reason is that relying solely on evaluating a single ranked response does not yield a stable measure. Instead, it is imperative to sample a fraction of responses to mitigate the variance and obtain a more accurate assessment, which is achieved by the proposed RETA metric.

Suppose instead of considering a single best response, we calculate the average of the top 32 responses and plot a \textit{best-32-of-n} curve (Figure~\ref{fig:BON vs RETA}(d)). When $n=128$, the value of such curve can be considered as an estimation of RETA$(1/4)$ without the normalizer (denominator $\mathbb{E}_{a}[{J}_{q}(a)]$ in Eq.~\eqref{equ:reta}), since we are evaluating the top 32 out of 128 responses. 
A closer inspect on the rankings of RMs, as indicated by the cutoff line in Figure~\ref{fig:BON vs RETA}(d) compared to the estimated RETA metric in Figure~\ref{fig:BON vs RETA}(a), shows that they are indeed very similar.
The slight discrepancies (e.g., RM5H-7B ranks differently) are caused by the absence of the normalizer.
We reckon that the normalizer is significant, as evaluation without it is not calibrated and is susceptible to prompt selection bias. The significance of the normalizer is further demonstrated in the ablation studies in Section~\ref{sec:ppl}.

\subsection{RETA Curve}
\begin{figure}[ht]
    \centering
    \vspace{-5mm}
    \includegraphics[width=9.5cm]{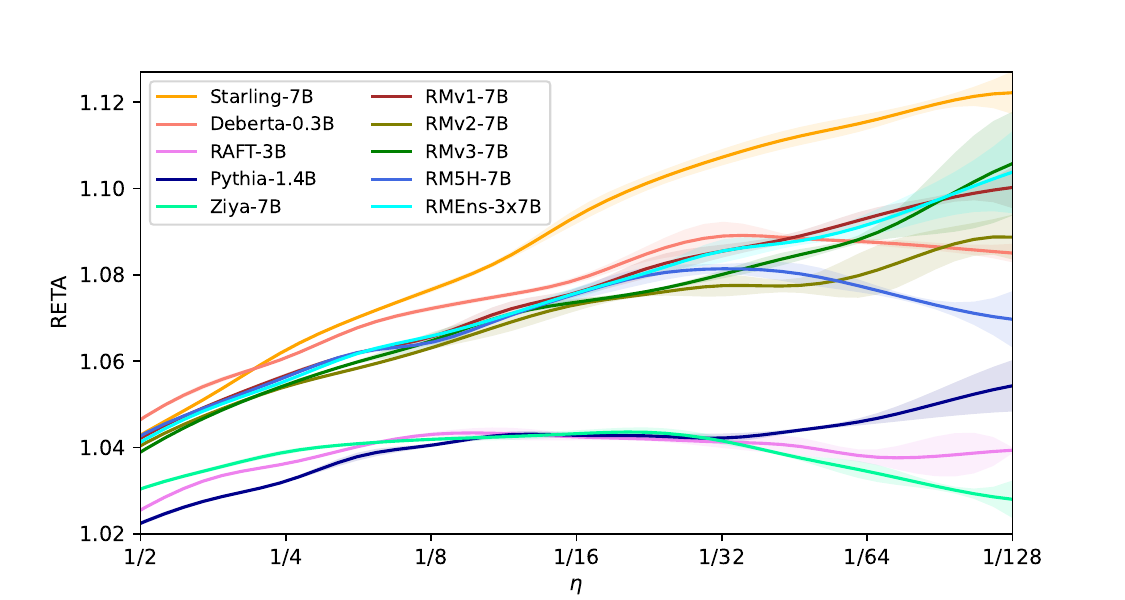}
    \vspace{-2mm}
    \caption{The RETA curves on Reliability-on-Helpfulness benchmark. The x-axis is plotted on a logarithmic scale ($-\log_2\eta$) for better visualization. Each curve is composed of 15 points connected by interpolation. The standard error across prompts is plotted. Note that as $\eta$ decreases, the evaluation tends to become noisier.}
    \label{fig:RETA curve}
\end{figure}
We present the RETA curves in Figure~\ref{fig:RETA curve}. For a RETA curve, each y-value represents the limit (Eq.~\eqref{equ:limit}) on the corresponding $\eta$ (x-value). Each y-value is estimated using the proposed method (as illustrated in Sec.~\ref{subsec: estimation} and Figure\ref{fig:BON vs RETA}(a)).
The RETA curve provides valuable insights as it characterizes the reliability of the RM at any quantile value. Ideally, as $\eta$ approaches zero, a reliable RM will select higher quality responses, thus we expect to observe an increasing trend in the curve. By analyzing the RETA curve, we can easily identify when an RM starts to become less reliable. For example, the performance of Deberta-0.3B, RM5H-7B, and Ziya-7B begins to decline as $\eta$ approaches zero. 
When using these RMs to select responses, we know the optimal quantile from which to select them.

There is additional information that can be observed from the RETA curve. It can be observed that Starling-7B consistently performs well across a wide range of $\eta$, indicating its overall reliability. However, it is surpassed by Deberta-0.3B around $\eta>1/3$. This suggests that the small model Deberta-0.3B exhibits superior reliability when selecting a large quantile of good responses. However, its parameter limitations might prevent it from effectively distinguishing the top-ranked responses.

The multi-head model RM5H-7B tends to exhibit higher reliability compared to its single-head counterpart, RMv3-7B, for $\eta\geq1/32$. However, RM5H-7B performs sub-optimally compared to RMv3-7B for $\eta<1/32$, suggesting that it might be more susceptible to the reward hacking issue in those cases.
In addition, we find that RMEns-3x7B, created through the ensemble method, appears to mitigate the reward hacking issue. Despite being a combination of three different RMs, RMEns-3x7B consistently performs close to the best performing RM among the three. This suggests that an ensemble of three equally competent RMs has the potential to further improve reliability.


\subsection{Ablation Study: Impact of the Normalizer}
\begin{figure}[ht]
\vspace{-3mm}
    \centering
    \includegraphics[width=12.4cm]{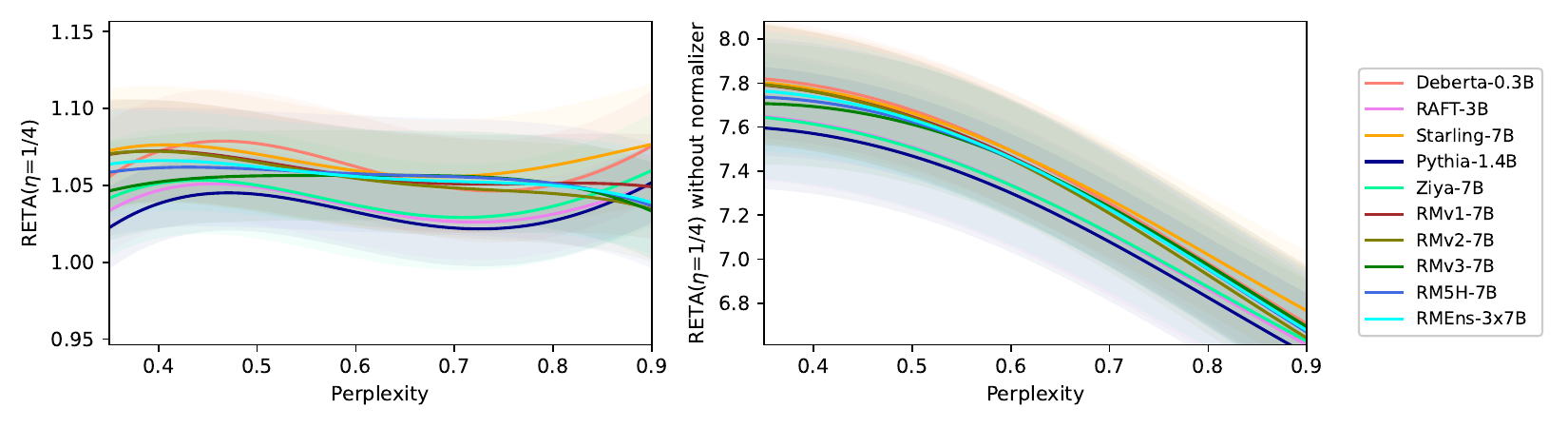}\hspace{-5mm}
    \\
    \vspace{-1mm}
    \hspace{-12mm}
    (a)\hspace{47mm}(b)
    \\
    \caption{The fitted curve of RETA metric versus prompt perplexity on Reliability-on-Helpfulness benchmark,
    (a) with normalizer,
    (b) without normalizer.
    The fitting is performed using Gaussian Processes, and curves in both figures utilize the same RBF kernel with a length scale of 0.5. To ensure a fair visual comparison, both figures plot the y-range as $\pm10\%$ of the mean value (of all curves).
    } 
    \label{fig:normalizer}
\end{figure}
\label{sec:ppl}

The normalizer is essential to reduce the influence of the choice of prompts. To demonstrate this, we characterize and compute the prompt perplexity as $\frac{1}{|\mathcal{A}_q|}\sum_{a\in\mathcal{A}_q}\text{PPL}_{\theta}(a|q)$ for each prompt $q$, and examine how RETA responds to prompt perplexity with and without the normalizer. As shown in Figure~\ref{fig:normalizer}, without the normalizer, the metric is sensitive to perplexity and exhibits a decreasing trend. With the normalizer, the metric shows resistance to perplexity, and the resulting curve appears more flat. Therefore, 
the normalizer can reduce the impact of potential prompt selection bias.
\subsection{Other Potential Metrics for Reliability Evaluation}
\begin{table}[ht]
\caption{The evaluation of various metrics on Anthropic-Helpful / Reliability-on-Helpfulness dataset. The top 3 RMs in each metric are highlighted in \textbf{bold}, while the worst 3 are \underline{underlined}. Higher values indicate better performance for all metrics.}
\centering
\scalebox{0.7}{
\begin{tabular}{c|ccc|ccc|cc|c|c}
\hline
 &  \multicolumn{3}{c|}{Accuracy ($\%$)} & \multicolumn{3}{c|}{Hit Rate ($\%$)} & \multicolumn{2}{c|}{Ranking} & {RETA($1/4$)} & Top 64 of 256 \\
Model & Test & TrainGen & TestGenPair & HR@32 & HR@64 & HR@128 & NDCG & MRR & Score & Win Rate ($\%$)  \\ \hline
Oracle       & 100            & 100            & 100            & 50             & 100            & 100            & 1.0             & 1.0             & 1.1840  & 52.93\\ \hline
Starling-7B  & \underline{70.64} & \textbf{77.67} & \textbf{77.38} & \textbf{25.23} & \textbf{45.99} & \textbf{78.64} & \textbf{0.2894} & \textbf{0.1383} & \textbf{1.0626}  & \textbf{32.88}\\
Deberta-0.3B & 71.68          & 72.91          & \textbf{76.68} & 22.25          & 42.14          & \textbf{76.75} & \underline{0.1996} & 0.0549          & \textbf{1.0608}  & \textbf{30.61}\\
RAFT-3B      & \textbf{83.34} & \underline{60.80} & \underline{42.77} & \underline{18.48} & \underline{36.50} & \underline{67.84} & \underline{0.1896} & \underline{0.0465} & \underline{1.0363}  & \underline{27.26}\\
Pythia-1.4B  & \underline{63.21} & \underline{63.01} & \underline{42.48} & \underline{19.07} & \underline{35.86} & \underline{66.70} & 0.2042          & 0.0580          & \underline{1.0322}  & \underline{27.09}\\
Ziya-7B      & \underline{65.72} & 68.92          & 72.01          & \underline{18.21} & \underline{36.25} & \underline{69.64} & \underline{0.1928} & \underline{0.0531} & \underline{1.0390}  & \underline{27.64}\\
RMv1-7B      & \textbf{74.95} & \textbf{73.52} & 73.30          & 22.78          & \textbf{42.36} & 74.59          & \textbf{0.2648} & \textbf{0.1208} & \textbf{1.0567}  & \textbf{30.78}\\
RMv2-7B      & 74.69          & 72.60          & \underline{71.64} & 22.33          & 41.75          & 73.67          & 0.2289          & 0.0802          & 1.0541  & 30.45\\
RMv3-7B      & 71.46          & \underline{68.49} & 73.82          & \textbf{23.04} & 42.22          & 73.93          & 0.2568          & 0.1060          & 1.0545  & 30.01\\ 
RM5H-7B      & 74.32          & 69.28          & \textbf{76.20} & 22.48          & 42.02          & \textbf{75.21} & 0.2003          & \underline{0.0506} & 1.0563  & 29.95\\ 
RMEns-3x7B   & \textbf{74.84} & \textbf{73.44} & 73.55          & \textbf{23.34} & \textbf{42.30} & 74.45          & \textbf{0.2692} & \textbf{0.1224} & 1.0555  & 30.51\\ \hline
        \end{tabular}}
\label{tab:metrics}
\end{table}

Besides BON and RETA, we also explore other potential metrics for evaluating the reliability of the RM. The results are presented in Table~\ref{tab:metrics}. The complete introduction of all metrics and related analyses are presented in Appendix~\ref{sec:other metrics}.

The first category is \textbf{Accuracy}, e.g. the original ``Test'' accuracy on the datasets used to build our benchmark, and two other accuracy metrics ``TrainGen'' and ``TestGenPair'' evaluating the preference of the generated response pairs. Our findings show that test metric alone is not sufficient to assess reliability. Despite RAFT-3B achieving the highest test accuracy ($83.34\%$) on Anthropic-Helpful and significantly surpasses the second most accurate approach ($74.95\%$), it is deemed unsatisfactory under all other reliability evaluation metrics. This suggests that suspicious data leakage cannot be identified by test accuracy alone. Therefore, relying solely on the accuracy metric and relevant accuracy benchmarks \cite{rewardbench} cannot provide comprehensive assessment of reliability.

The well-known ranking metrics, including the \textbf{Hit Rate}, \textit{Normalized Discounted Cumulative Gain} (\textbf{NDCG}) \cite{ndcg} and \textit{Mean Reciprocal Ranking} (\textbf{MRR}),
all have specific limitations in assessing the reliability of RM (Appendix~\ref{sec:other metrics}).
In comparison, the proposed RETA metric provides convincing assessments as its results align well with the sampled ``Win Rate'' against GPT4.

\section{Conclusions and Future Work}
In this paper, we introduce the RETA metric as a principled approach for evaluating the reliability of RMs. We highlight its advantages and establish an integrated benchmarking pipeline based on it. Through extensive experimental studies, we obtain compelling and valuable insights into the reliability of various publicly available as well as proprietary RMs.

A promising direction to extend this study is to analyze the relationship between reliability and performance of RMs in terms of guiding RLHF training for LLMs. Exploring this connection would require extensive experiments to obtain compelling results, which we plan to conduct in subsequent studies.
Another potential avenue for future research is to train RMs (or develop techniques specifically designed) to maximize the RETA metric directly. By analyzing the effectiveness of the resulting RMs, we can assess whether overall reliability is indeed improved.

\newpage

\bibliographystyle{plain}
\bibliography{ref}

\newpage
\appendix

\section{More Experimental Results}
\subsection{Other Potential Metrics for Reliability Evaluation}
\label{sec:other metrics}
We also explore other potential metrics for evaluating the reliability of the RM. The first category we consider is \textbf{accuracy}, which includes the following metrics:
(1) Test accuracy (referred to as \textbf{Test}): This metric measures the accuracy of the RM on the test split of the Anthropic-Helpful dataset. 
(2) Train generation accuracy: This metric evaluates the accuracy of comparing a pair of responses randomly generated using the Llama model for prompts in the train split of the Anthropic-Helpful dataset. The ground truth preference is determined by our oracle (GPT4). This metric is abbreviated as \textbf{TrainGen}. 
(3) Test generation pairwise accuracy: Similar to TrainGen, but this metric assesses the preference between all possible pairwise responses from our curated response set, which concerning prompts sampled from the test split. This metric is abbreviated as \textbf{TestGenPair}.

We also incorporate some well-known ranking metrics, as they allow us to directly assess the quality of the ranked list of responses generated by the RM. 
Firstly the \textbf{Hit Rate}, we define the top $\eta=25\%$ quantile of the oracle's selected responses (a total of 64 responses per prompt) as the groundtruth set with positive labels. By setting such a specific cutoff range, we can calculate the hit rate, which measures the proportion of correctly identified positive responses.
Additionally, we include \textit{Normalized Discounted Cumulative Gain} (\textbf{NDCG}) \cite{ndcg} and \textit{Mean Reciprocal Ranking} (\textbf{MRR}). These metrics focus on evaluating the ranking of the best answer selected by the RM.

In addition, we introduce a metric, the \textbf{Win Rate} against GPT4, as a ground truth evaluation of the quality of the responses selected by the RM. We select the top 64 (out of 256) responses rated by the RM, and for each of them, we compare them with GPT4's answer and ask GPT4 to determine which answer is better (or if they are a draw).

\begin{figure}[ht]
\vspace{-5mm}
    \centering
    \subfigure[\hspace{-2mm}]{\includegraphics[width=0.32\textwidth]{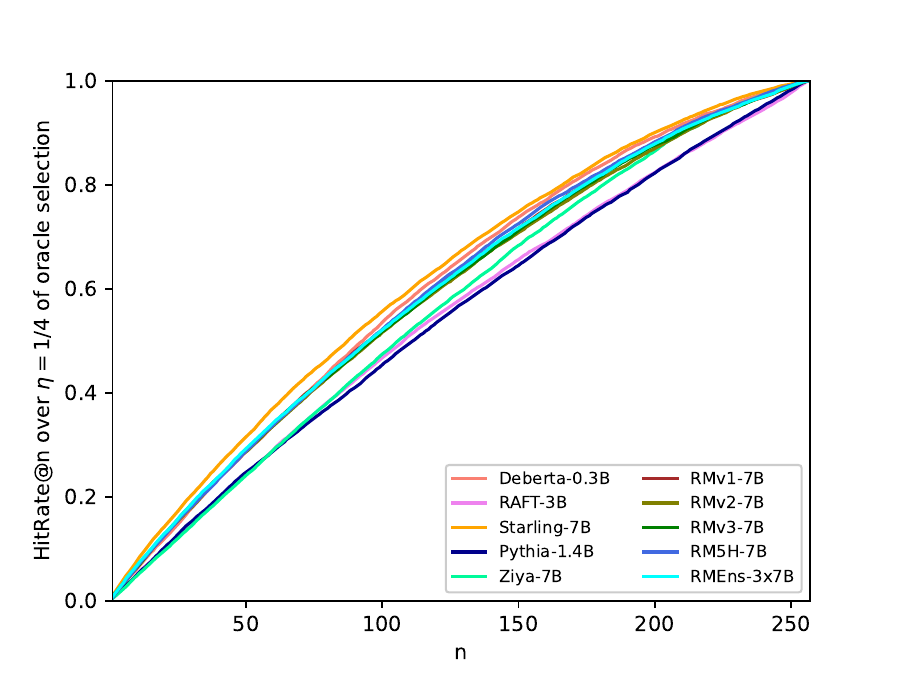}} 
    \subfigure[\hspace{-2mm}]{\includegraphics[width=0.32\textwidth]{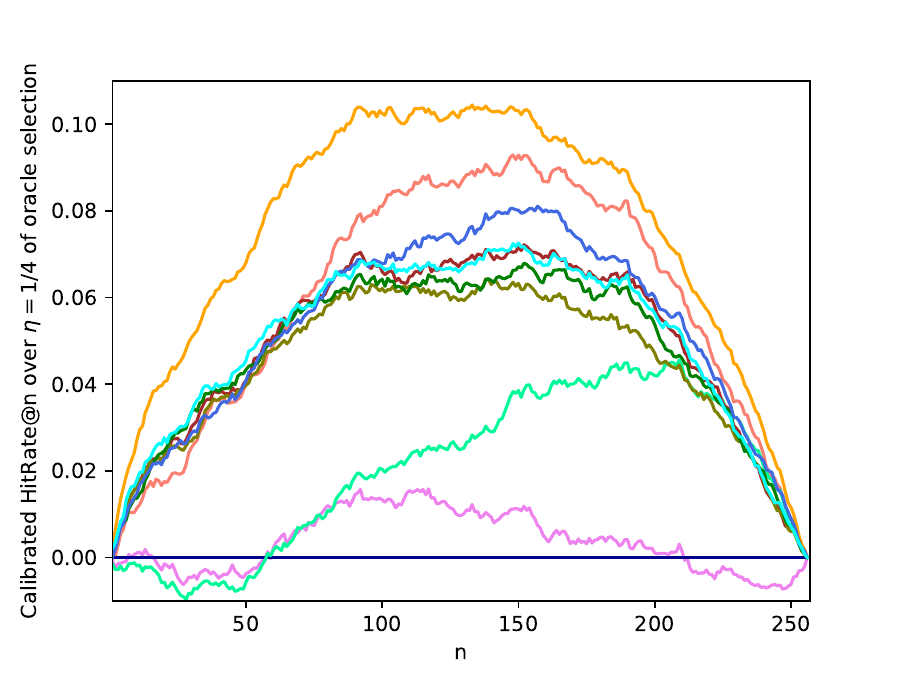}} 
    \subfigure[\hspace{-2mm}]{\includegraphics[width=0.32\textwidth]{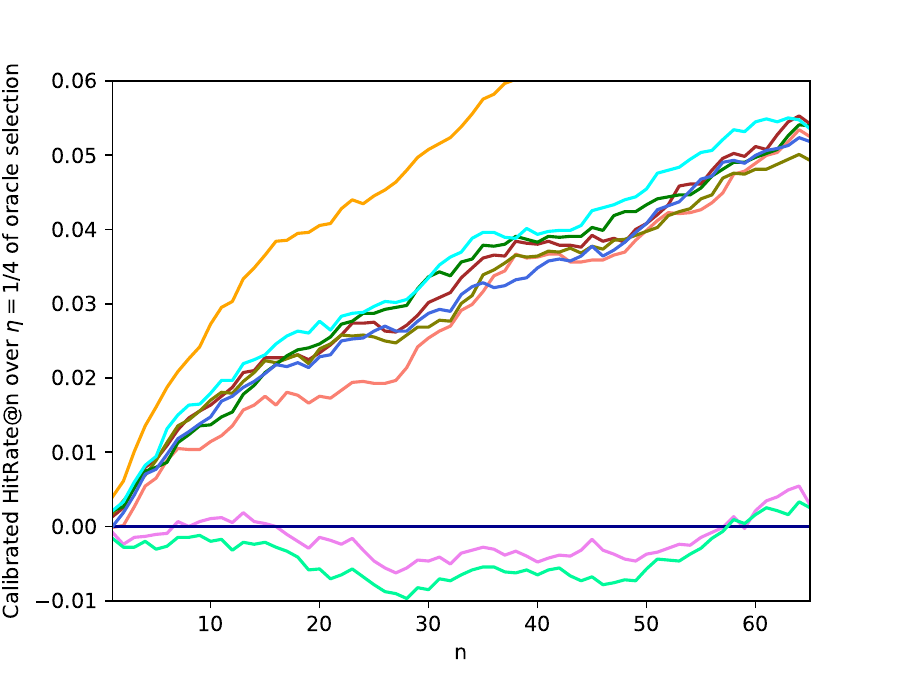}}
    \vspace{-1mm}
    \caption{The \textbf{Hit Rate} metric evaluated on Reliability-on-Helpfulness dataset: 
    (a) The hit rate at $n$ is calculated by comparing the top $n$ responses generated by the RM with the ground truth set, which is defined as the $\eta=1/4$ quantile of responses selected by the oracle. 
    (b) The replot of (a) by subtracting the result of Pythia-1.4B as the baseline.
    (c) The zoom-in version of (b) which provides a closer look at the head region.
    } 
    \label{fig:hitrate eta 1o4}
\end{figure}

The results are reported in Table~\ref{tab:metrics} of the main text. From the table, we can further make the following observations:
\begin{itemize}[leftmargin=10pt]
    \item     The \textbf{Test} metric alone is not sufficient to imply reliability. Despite RAFT-3B achieving the highest test accuracy and a significant lead over the second-place, it is deemed unsatisfactory in all other evaluations. This suggests that there may be data leakage in its training process, which cannot be determined by test accuracy alone. Therefore, relying solely on the \textbf{Test} metric and relevant accuracy benchmarks \cite{rewardbench} could not provide a comprehensive assessment of reliability.
    \item     The \textbf{TrainGen} metric appears to be acceptable, but it comes with a high labeling cost. In the evaluation in Table~\ref{tab:metrics}, we did not perform subsampling, and the total labeling cost (for two responses per prompt) is 1.68 times the cost of building our Reliability-on-Helpfulness benchmark. 
    \item     The \textbf{TestGenPair} metric is not trustworthy. We observe that Ziya-7B, despite being one of the least reliable RMs in other evaluations, performs relatively well according to this metric. One possible explanation is that this pairwise accuracy, being a global metric, may not effectively capture the specific areas where the RM reckons as good, which is more important in assessing the reliability.
    \item     We further explored the \textbf{Hit Rate} and analyzed the results presented in Figure~\ref{fig:hitrate eta 1o4}. It appears that the hit rate metric shows less differentiation among the different RMs. Additionally, a well-known limitation of hit rate metric is that it does not take into account the rankings within the groundtruth set defined through oracle.
    \item     The ranking metrics, \textbf{NDCG} and \textbf{MRR}, which solely focus on the best answer selected for each prompt, suffer from the same issue of instability as the BON curve evaluation. The results presented in Table~\ref{tab:more on ndcg and mrr} demonstrate that while NDCG and MRR themselves exhibit high consistency in evaluation, their assessment of RMs changes drastically when we shift our attention from the best answer selected by the RM to the second best answer selected. The instability makes them unreliable since they fail to provide consistent assessments of the RMs' performance.
    \item      The proposed RETA metric provides convincing assessments as its results align well with the sampled \textbf{Win Rate} against GPT4. It highlights the soundness of the RETA metric. If we delve deeper into the evaluation process, we can consider the win/lose/draw outcome as a direct oracle output, assigning values such as 1 for a win, 0 for a draw, and -1 for a loss. Therefore, the \textbf{Win Rate} can be seen as a variant of the RETA metric using this oracle labeling scheme. This observation of high consistency indicates that the evaluation of the RETA metric is somewhat robust to the choice of oracle labeling scheme.
    \item     The ensemble method RMEns-3x7B performs well across various metrics, suggesting that it is an effective approach for enhancing reliability.
\end{itemize}

\begin{table}[ht]
\caption{The evaluation of two ranking metrics on Reliability-on-Helpfulness dataset. The top 3 RMs in each metric are highlighted in \textbf{bold}. Higher values indicate better performance for all metrics.}
\centering
\scalebox{0.78}{
\begin{tabular}{c|cc|cc}
\hline
             &  \multicolumn{2}{c|}{Evaluating the best response}                               & \multicolumn{2}{c}{Evaluating the second best response} \\
Model        & NDCG                            & MRR                             & NDCG            & MRR \\ \hline
Oracle       & 1.0                             & 1.0                             & 0.7970                          & 0.725                 \\ \hline
Starling-7B  & \textbf{0.2894} (1st out of 10) & \textbf{0.1383} (1st out of 10) & \textbf{0.2697} (1st out of 10) & \textbf{0.1240} (1st out of 10) \\
Deberta-0.3B & 0.1996          (8th out of 10) & 0.0549          (7th out of 10) & 0.2153          (7th out of 10) & 0.0691 (7th out of 10) \\
RAFT-3B      & 0.1896         (10th out of 10) & 0.0465         (10th out of 10) & 0.1745          (9th out of 10) & 0.0317 (9th out of 10) \\
Pythia-1.4B  & 0.2042          (6th out of 10) & 0.0580          (6th out of 10) & 0.1873          (8th out of 10) & 0.0421 (8th out of 10) \\
Ziya-7B      & 0.1928          (9th out of 10) & 0.0531          (8th out of 10) & 0.1708         (10th out of 10) & 0.0261 (10th out of 10) \\
RMv1-7B      & \textbf{0.2648} (3rd out of 10) & \textbf{0.1208} (3rd out of 10) & 0.2331          (4th out of 10) & 0.0870 (4th out of 10) \\
RMv2-7B      & 0.2289          (5th out of 10) & 0.0802          (5th out of 10) & \textbf{0.2382} (3rd out of 10) & \textbf{0.0884} (3rd out of 10) \\
RMv3-7B      & 0.2568          (4th out of 10) & 0.1060          (4th out of 10) & 0.2241          (5th out of 10) & 0.0779 (5th out of 10) \\ 
RM5H-7B      & 0.2003          (7th out of 10) & 0.0506          (9th out of 10) & \textbf{0.2394} (2nd out of 10) & \textbf{0.0939} (2nd out of 10) \\ 
RMEns-3x7B   & \textbf{0.2692} (2nd out of 10) & \textbf{0.1224} (2nd out of 10) & 0.2193          (6th out of 10) & 0.0719 (6th out of 10) \\ \hline
\end{tabular}}
\label{tab:more on ndcg and mrr}
\end{table}

\subsection{Our Second Benchmark: Reliability-on-Multi-Turn-Conversation}
\label{sec:benchmark2}
\textbf{Reliability-on-Multi-Turn-Conversation.} 
Many existing datasets only measure LLMs’ core capabilities on a limited set of tasks, failing to adequately assess their alignment (such as accurately adhering to instructions) in multi-turn dialogues.
MT-bench \cite{mtbench} addresses this gap by providing a series of $|\mathcal{Q}_0|=80$ open-ended prompt sets designed to evaluate a chatbot’s multi-turn conversational and instruction-following abilities. It consists of carefully constructed prompt sets that differentiate chatbots based on their core capabilities, such as reasoning and math. Each prompt set includes two turns of prompts, so we concatenate the embeddings of these two prompts to create the prompt set embedding for DPP. We subsampled $k=20$ prompt sets and correspondingly obtained $N=256$ RP response sets (each has two-turn responses). Our oracle evaluation follows the suggested format of \cite{mtbench} to present two full conversations in a single evaluation template in which we ask the oracle to moderately focus on the second conversation. The total labeling cost for this benchmark is $\$64.0$.

\begin{figure}[ht]
\vspace{-5mm}
    \centering
    \subfigure[\hspace{-7mm}]{\includegraphics[width=0.335\textwidth]{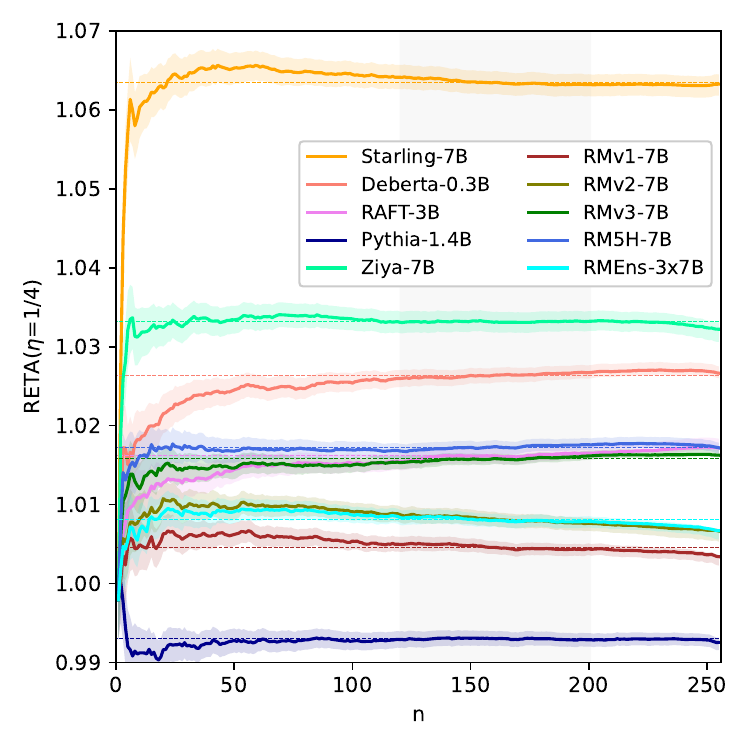}}
    \hspace{-3mm}
    \subfigure[\hspace{-7mm}]{\includegraphics[width=0.33\textwidth]{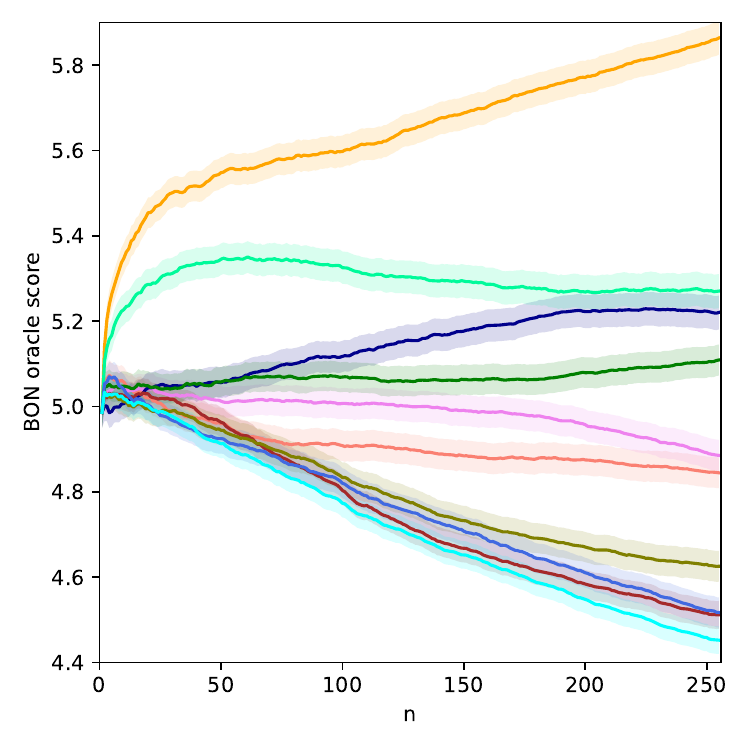}} 
    \hspace{-3mm}
    \subfigure[\hspace{-7mm}]{\includegraphics[width=0.33\textwidth]{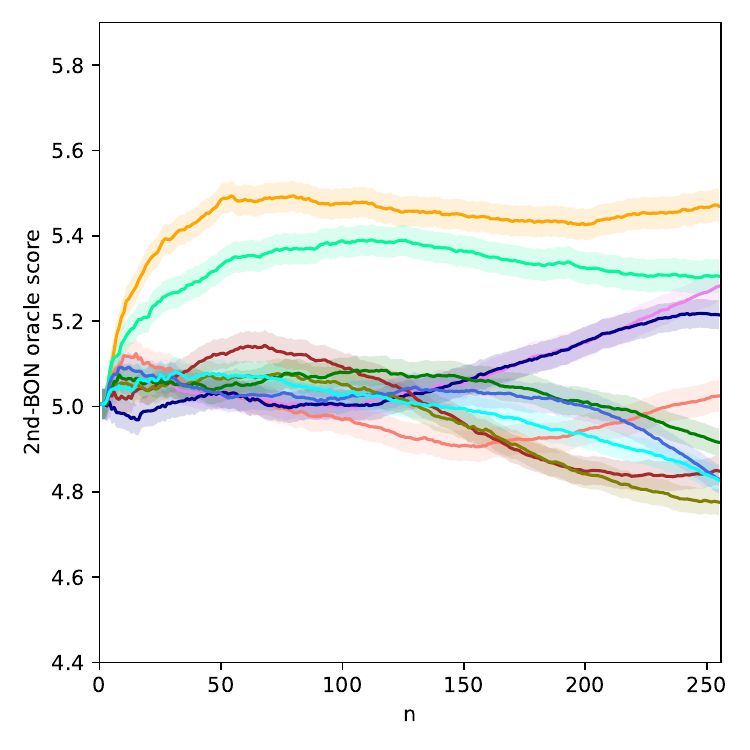}}
    \vspace{-1mm}
    \caption{The results on Reliability-on-Multi-Turn-Conversation benchmark: 
    (a) The estimation of RETA($\eta=1/4$) in Eq.~\eqref{equ:approx} versus resampled size $n$. 
    Dashed horizontal lines mark the limiting values, with the light grey shaded area representing the range used to calculate the final estimation of RETA's limiting values (Sec.~\ref{subsec: estimation}).
    (b) \textit{Best-of-n} (BON) curve versus $n$. The x-axis can also be transformed into KL divergence without loss of generality. 
    (c) \textit{2nd-best-of-n} curve versus $n$.
    } 
    \label{fig:mtbench1}
\end{figure}

\begin{figure}[ht]
    \centering
    \vspace{-5mm}
    \includegraphics[width=9.5cm]{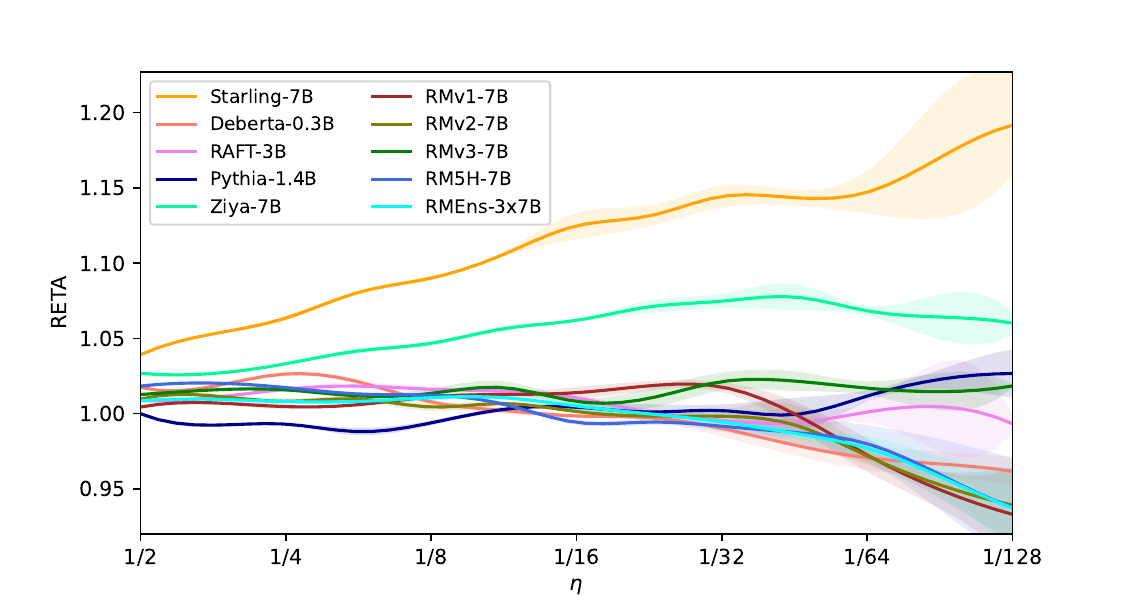}
    \vspace{-2mm}
    \caption{The RETA curves on Reliability-on-Multi-Turn-Conversation benchmark. The x-axis is plotted on a logarithmic scale ($-\log_2\eta$) for better visualization. Each curve is composed of 15 points connected by interpolation. The standard error across prompts is plotted.}
    \label{fig:mtbench2}
\end{figure}

The following observations can be made: 

\begin{itemize}[leftmargin=10pt]
\item From Fig.~\ref{fig:mtbench1}, similar to the Reliability-on-Helpfulness benchmark, RETA achieves rapid convergence with respect to $n$ and converges to a desirable limiting value. The instability at the tail end can occasionally be seen. It again underscores the necessity of our proposed estimation scheme.
\item From Fig.~\ref{fig:mtbench1} and Fig.~\ref{fig:mtbench2} we observe that Starling-7B continues to demonstrate the highest performance, while Ziya-7B, a capable model on MT-bench as tested in \cite{rewardbench}, also exhibits strong reliability and secures the second position among the tested RMs on our Reliability-on-Multi-Turn-Conversation benchmark.
\item The Reliability-on-Multi-Turn-Conversation benchmark is more challenging than the Reliability-on-Helpfulness benchmark. This is evident from Figure \ref{fig:mtbench2}, where we observe that several RMs perform at the level of a random baseline, and some even perform worse than random.
\item The ensembling approach (RMEns-3x7B) and multi-head structures (RM5H-7B) appear to be ineffective in improving reliability performance in this benchmark.
\end{itemize}

\section{Proof, Training Details, and Benchmarking Details}
\subsection{The Proof of the Limiting Value of RETA}
\label{appendix:limit}

Before delving into the formal proof of the main theorem, we first introduce a lemma that will be utilized later. Additionally, we introduce some notations here. Let $X_{(j:n)}$ denote the $j$-th largest datum among $X_1,\ldots,X_n$.
It is worth noting that we slightly abuse the notation by using $a$ to represent both the variable and its observations.

\begin{lemma}
    If $X$ is a random variable with distribution function $F$, and for $0<\eta<1$ define the quantile function as $\Theta(\eta)=\inf (x:F(x)\geq 1-\eta)$. Then $X_n\rightarrow X$ in distribution if and only if $\Theta_n(\eta)\rightarrow \Theta(\eta)$ at all continuity points $\eta$ of $\Theta$.
\end{lemma}
This lemma is the result of Proposition 5 in \cite{fristedt2013modern}, and is also proved in Chapter 21 of \cite{van2000asymptotic}. Here $\Theta_n(\eta)$ means the $\eta$ quantile of samples $X_n$, i.e. $\Theta_n(\eta)=X_{(\eta\times n:n)}$ in the case that $\eta\times n$ is an integer.

\begin{proof}
We can deduce
\begin{align*}
&
\lim_{|\mathcal{A}_q|\to\infty}\;\;\frac{1}{\eta\times |\mathcal{A}_q|}\mathbb{E}_{\mathcal{A}_q}\left[
\sum_{a\in\text{BETA}_{Y}(\mathcal{A}_q,\eta)} {J}_{q}(a)\right]
\\
=&\lim_{n\to\infty}\;\;\frac{1}{\eta\times n}\sum_{j=1}^{\eta\times n} \mathbb{E}_{a}\left[{
{J}_{q}(a)|{X_{(j-1:n)}\geq Y_q(a)\geq X_{(j+1:n)}}}\right]
\\
=&\lim_{n\to\infty}\;\;\frac{1}{\eta\times n}\sum_{j=1}^{\eta\times n} \int
{J}_{q}(a)\;p(a|{X_{(j-1:n)}\geq Y_q(a)\geq X_{(j+1:n)}})\;da
\\
=&\lim_{n\to\infty}\;\;\frac{1}{\eta}\sum_{j=1}^{\eta\times n} \int
{J}_{q}(a)\;p({X_{(j-1:n)}\geq Y_q(a)\geq X_{(j+1:n)}}|a)\;p(a)\;da
\\
=&\lim_{n\to\infty}\;\;\frac{1}{\eta} \int
{J}_{q}(a)\;p(Y_q(a)\geq X_{(\eta\times n:n)}|a)\;p(a)\;da
\\
=&\frac{1}{\eta} \int
{J}_{q}(a)\;p(Y_q(a)\geq \Theta(\eta)|a)\;p(a)\;da
\end{align*}
The first equality holds due to the linearity of expectation, where the conditional expectation $X_{(j-1:n)}\geq Y_q(a)\geq X_{(j+1:n)}$ represents the condition that $Y_q(a)$ happens to be the $j$-th largest datum of $X_1,\ldots,X_n$; The third equality follows from Bayes' rule, where $p(X_{(j-1:n)}\geq Y_q(a)\geq X_{(j+1:n)})=1/n$ for an arbitrary $a$; The forth equality follows from summation of cases; The last equality follows from the above Lemma about convergence of sample quantile. 

Now, given the assumption that ${J}_{q}$ is bounded, we have successfully demonstrated the validity of the RETA metric's limit. To simplify the expression further, we can apply Bayes' rule.
\begin{align*}
&\frac{1}{\eta} \int
{J}_{q}(a)\;p(Y_q(a)\geq \Theta(\eta)|a)\;p(a)\;da
\\
=&\frac{1}{\eta} \int
{J}_{q}(a)\;p(a|Y_q(a)\geq \Theta(\eta))\;p(Y_q(a)\geq \Theta(\eta))\;da
\\
=&\int
{J}_{q}(a)\;p(a|Y_q(a)\geq \Theta(\eta))\;da
\\
=& \mathbb{E}_{a}\left[{
{J}_{q}(a)|{Y_q(a)\geq \Theta(\eta)}}\right].
\end{align*}
With the aforementioned results and the expression of the RETA metric, we can conclude and complete the rest of the proof.
\end{proof}


\subsection{Smoothing Techniques and the Final Estimator}
\label{appendix:smoothing}
In situations where $\eta\times n$ is not an integer, we still aim to utilize it as an intermediate value to reduce the variance in our final estimation of RETA. To address this, we propose the use of the following asymptotic unbiased estimator. Firstly, we refine our definition of the BETA subset to accommodate non-integer cases:
\[
\textstyle
\text{BETA}_{Y}(\mathcal{A}_q,\eta)\equiv
\argmax_{\mathcal{A} \subset \mathcal{A}_q: |\mathcal{A}| =\left\lfloor\eta\times |\mathcal{A}_q|\right\rfloor} \sum_{a \in \mathcal{A}} Y_{q}(a),
\]
Here $\lfloor x \rfloor$ denotes the floor function (i.e., for any real number $x$ it returns the largest integer $y$ such that $y \leq x$). To simplify the following notations, we introduce the residual function $\Delta(x)\equiv x-\lfloor x \rfloor$. Additionally, we let $a_{(j:n)}$ denotes the specific responses among $\mathcal{A}={a_{i_1},\ldots,a_{i_n}}$, such that $Y_q(a_{(j:n)})=X_{(j:n)}$, where $X_{(j:n)}$ is the $j$-th largest datum among $X_1,\ldots,X_n$. With these notations, we can now present our asymptotic unbiased estimator, which substitutes Equation~\eqref{equ:approx} in the main text when $\eta\times n$ is not an integer and is equivalent to Equation~\eqref{equ:approx} when $\eta\times n$ is an integer:
\begin{equation}
\begin{split}
\textstyle
\frac{1}{|\mathcal{Q}|}\sum_{q\in\mathcal{Q}}\Big[
\frac{N}{\eta n}\mathbb{E}_{\mathcal{A}}
\Big[&\sum_{a\in\text{BETA}_{Y}(\mathcal{A},\eta)} {J}_{q}(a)+
\Delta(\eta n)\cdot
\\&
\big[\Delta(\eta n)\cdot{J}_{q}(a_{(\lfloor \eta n \rfloor+1:n)})+(1-\Delta(\eta n))\cdot{J}_{q}(a_{(\lfloor \eta n \rfloor:n)})\big]\Big]
/{\sum_{a\in\mathcal{A}_q}{J}_{q}(a)}
\Big].
\label{equ:approx non-int}
\end{split}
\end{equation}
Note that the asymptotic unbiasedness can be easily understood as $a_{(\lfloor \eta n \rfloor:n)}$ and $a_{(\lfloor \eta n \rfloor+1:n)}$ will be arbitrarily close as $n$ approaches infinity.

\subsection{Efficient Unbiased Sampler for BON}
\label{appendix:bon sampler}
In theory, when employing a Monte Carlo estimator for the expectation mentioned above, the naive way will repeatedly sample disjoint sets of $\mathcal{A}_q$ for various values of $n$. This implies that the sampled responses are not reused, which is highly inefficient and wasteful. To address this, \cite{webgpt} initially proposed sampling $N\geq n$ responses and subsequently adopting the following provably unbiased estimator:
\[
\textstyle
\frac{1}{|\mathcal{Q}|}\sum_{q\in\mathcal{Q}}\left[
\frac{1}{\binom{N}{n}} \sum_{1\leq i_1\leq\ldots\leq i_n\leq N}{J}\left(q,\argmax_{a \in \{a_{i_1},\ldots,a_{i_n}\}} Y_{q}(a)\right)\right].
\]

Thus, the typical preparation procedure involves initially invoking the oracle to evaluate all $N$ sampled responses. Subsequent RM evaluations do not require additional oracle labeling. Consequently, the total oracle labeling cost amounts to $O(|\mathcal{Q}|N)$. This has emerged as the standard practice in evaluating the BON curve \cite{webgpt,gao2023scaling}.

\subsection{The Choice of Reference Policy}
\label{appendix:RP model}

In principle, the Reference Policy (RP) can be a pretrained LLM, an instruction-fine-tuned (IFT) LLM (e.g., as described in \cite{gao2023scaling}), or a LLM after the RLHF training. If the goal is to assess the reliability of the RM to guide the training of an RL policy, one promising approach could be to use the initial policy model as the RP. However, in this work, for generality, we have considered a RP model that is both competent and comprehensive.

In our study, we have found that the pretrained model Llama2-7B have difficulty in generating good responses, whereas Llama2-7B-Chat appeared to be more capable in this regard. Consequently, we selected Llama2-7B-Chat as our RP for the entirety of our paper.

While considering a mixture of diverse RP models to achieve a broader coverage of the input distribution could be advantageous, we leave this as a direction for future research.


\subsection{Evaluation with Oracle}
\label{appendix:oracle}
\begin{figure}[ht]
\vspace{-0mm}
    \centering\includegraphics[width=0.95\textwidth]{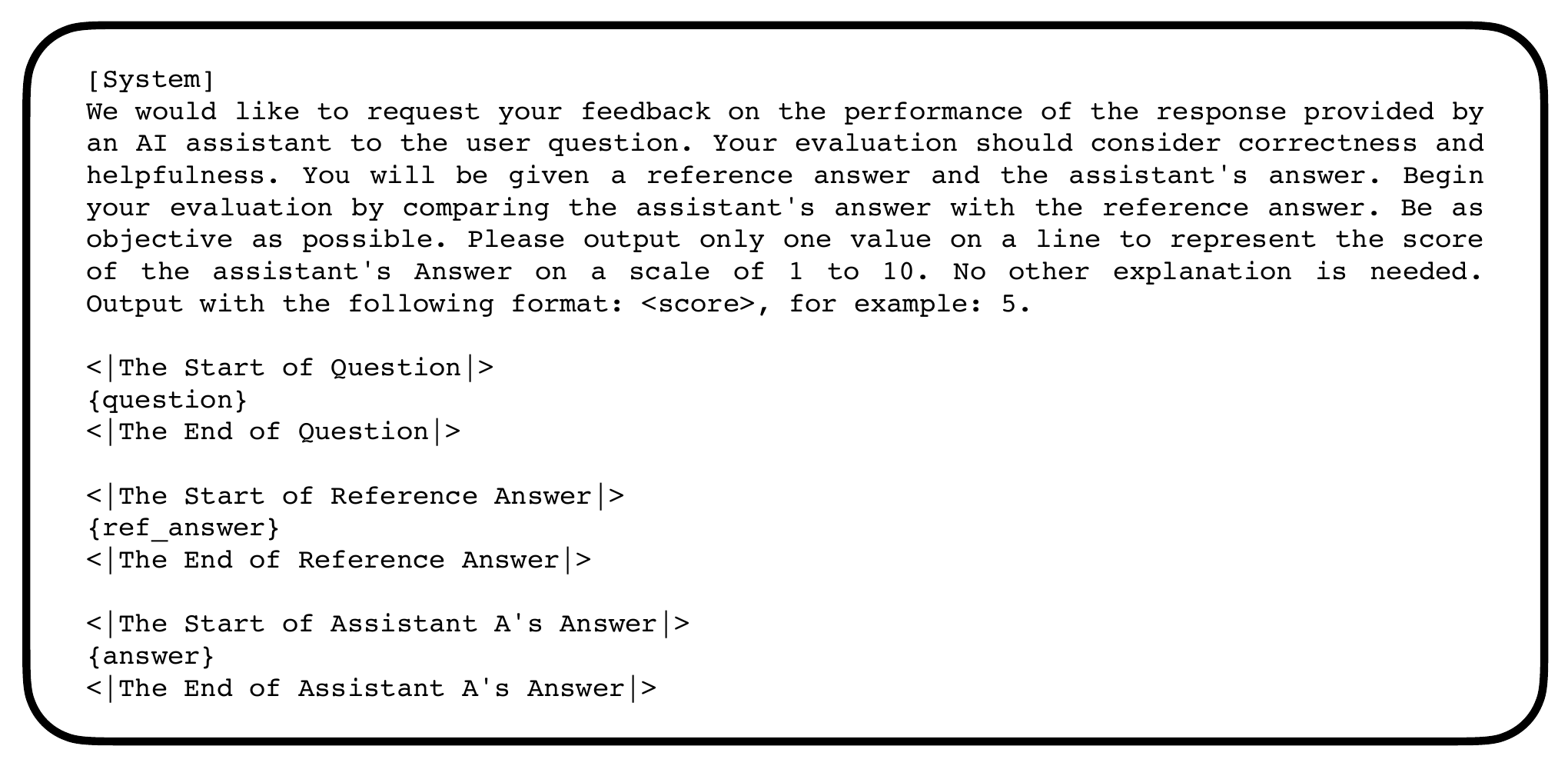}
    \vspace{-1mm}
    \caption{The prompt for helpfulness oracle evaluation when building the Reliability-on-Helpfulness benchmark.
    } 
    \label{fig:hh_template}
\end{figure}

\begin{figure}[ht]
\vspace{-0mm}
    \centering\includegraphics[width=0.95\textwidth]{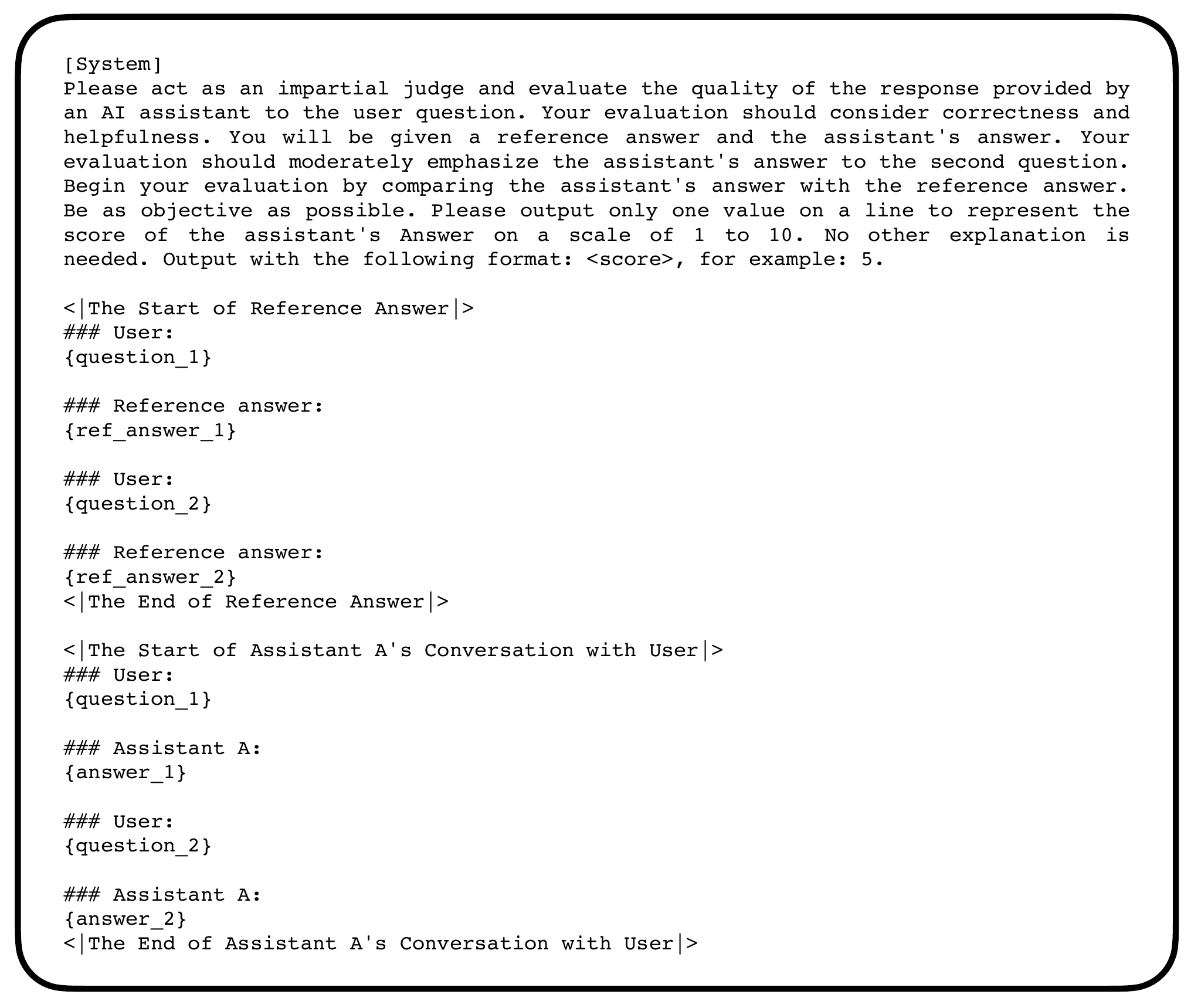}
    \vspace{-1mm}
    \caption{The prompt for multi-turn oracle evaluation when building the Reliability-on-Multi-Turn-Conversation benchmark. We adopt a similar fashion to the evaluation template used in \cite{mtbench}.
    } 
    \label{fig:mt_template}
\end{figure}

We provide the oracle evaluation templates for Reliability-on-Helpfulness and Reliability-on-Multi-Turn-Conversation benchmarks in Figure \ref{fig:hh_template} and Figure \ref{fig:mt_template}, respectively. It is important to note that for Reliability-on-Multi-Turn-Conversation benchmark, due to the limitations of the RM evaluation in terms of multi-turn conversation, we ask the oracle to moderately focus on the second question.

OpenAI's GPT4 API inherently supports multiple outputs in a single query. Therefore, we obtain 10 outputs and calculate their average as the ground truth score.

\subsection{Training Procedure for RMv1-7B, RMv2-7B, RMv3-7B and RM5H-7B}
\label{appendix:self train RMS}
Our reward models are initialized from the pretrained model Llama2-7B, leveraging the knowledge gained during pretraining for enhanced performance. In the case of RMv1-7B, RMv2-7B, and RMv3-7B, we adopt the standard practice of substituting the classification head, used for next-token prediction, with a regression head that outputs a scalar reward. The specific hyperparameters for each model vary and are detailed below.
The training dataset utilized is the training partition of the Anthropic-Helpful base dataset, which comprises approximately 43k preference pairs.

\begin{table}[ht]
\centering\caption{
The trained models RMv1-7B, RMv2-7B, and RMv3-7B have distinct hyperparameter values as outlined below. Aside from these specified hyperparameters, the models share common settings, including $1$ training epoch, a cosine learning rate schedule with a warm-up ratio of $0.03$, and a weight decay factor of $0.1$.}
\begin{tabular}{c|ccc}
\hline
Models  & Batch Size & Learning Rate & Max Length \\ \hline
RMv1-7B                               & 16         & 1.2e-5        & 1024       \\
RMv2-7B                               & 16         & 1.5e-5        & 2048       \\
RMv3-7B                               & 8          & 2e-5          & 2048       \\ \hline
\end{tabular}
\end{table}

Specifically, RM5H-7B shares identical hyperparameter settings with RMv3-7B, with the sole distinction being its multi-head architecture. Rather than utilizing a single regression head, RM5H-7B employs five heads, each producing a scalar reward value. The final reward is calculated by taking the average of these five values. To mitigate overfitting during training, we employ a strategy where only two out of the five heads are randomly updated at each training step.

\end{document}